\PassOptionsToPackage{table}{xcolor}
\documentclass[sigconf]{acmart}
\usepackage{booktabs}
\usepackage{graphicx}
\usepackage{makecell}
\usepackage{pifont}
\usepackage{multirow}
\usepackage{subcaption}
\definecolor{HeaderGray}{gray}{0.9}
\definecolor{myExtremelyLightBlue}{RGB}{235,245,255}
\definecolor{markgreen}{RGB}{0,128,70}
\definecolor{markred}{RGB}{190,40,40}

\definecolor{closedgray}{HTML}{777777}
\definecolor{overallcolor}{HTML}{00008B}

\newcommand{\closed}[1]{\textcolor{closedgray}{#1}}

\usepackage{xcolor}
\usepackage{graphicx}
\usepackage{wrapfig}
\usepackage[most]{tcolorbox}

\definecolor{correctgreen}{RGB}{0,150,80}

\newtcolorbox{ReasoningBox}[2][]{
    enhanced,
    breakable, 
    colback=white,
    colframe=black,
    coltitle=white,
    fonttitle=\bfseries,
    title={#2},
    sharp corners,
    boxrule=1pt,
    attach boxed title to top left={xshift=0mm, yshift=0mm},
    boxed title style={
        sharp corners, 
        colback=black, 
        colframe=black, 
        boxrule=1pt,
        right=2mm, left=2mm
    },
    top=1.5em,
    #1
}

\newcommand{\eat}[1]{}

\newcommand{\ours}{\textsc{ClinPRISM}}

\definecolor{myExtremelyLightGreen}{RGB}{232, 245, 233}
\definecolor{tablerowcolor}{RGB}{242,242,242}
\definecolor{tablerowcolor1}{RGB}{242,242,242}

\AtBeginDocument{%
  }

\setcopyright{acmlicensed}
\copyrightyear{2018}
\acmYear{2018}
\acmDOI{XXXXXXX.XXXXXXX}
\acmConference[Conference acronym 'XX]{Make sure to enter the correct
  conference title from your rights confirmation email}{June 03--05,
  2018}{Woodstock, NY}
\acmISBN{978-1-4503-XXXX-X/2018/06}

\begin{document}

\title{A Cost-Effective Multimodal LLM Reasoning Framework for Question Answering over Irregular Clinical Time Series}

\author{Frank Nie}
\authornote{Equal Contribution.}
\affiliation{%
  \institution{Shandong University}
  \country{China}
}

\author{Ethan B. Liu}
\authornotemark[1]
\authornote{Corresponding author.}
\affiliation{%
  \institution{Shandong University}
  \country{China}
}

\author{Yuan Zhu}
\authornotemark[1]
\affiliation{%
  \institution{Shandong University}
  \country{China}
}
\email{zhu948982@gmail.com}

\author{Wei Fan}
\affiliation{%
  \institution{University of Auckland}
  \country{New Zealand}
}

\author{Jindong Han}
\authornotemark[2]
\affiliation{%
  \institution{Shandong University}
  \country{China}
}
\email{jindong.han@sdu.edu.cn}


\renewcommand{\shortauthors}{Trovato et al.}

\begin{abstract}
Question answering (QA) over irregular clinical time series (ICTS) plays a pivotal role in a wide range of healthcare applications. Although recent multimodal time-series large language models (LLMs) have shown considerable promise in general-purpose time-series QA, they remain poorly equipped to model the sparsity, asynchrony, and irregular sampling patterns of clinical observations. To fill this gap, we propose \ours, a cost-effective multimodal LLM reasoning framework for question answering over ICTS data. First, we devise an irregularity-aware multi-scale encoder to capture sparse clinical evidence at diverse temporal scales. Then, we propose a temporal evidence distiller to integrate representations across these scales and compress them into a small number of LLM-compatible tokens. Moreover, we introduce a progressive alignment strategy that sequentially aligns the irregular trajectories with the LLM’s textual embedding space. To facilitate training, we construct 30,000 clinical time series paired with multi-scale descriptions, together with 41,000 instruction-tuning instances spanning 11 tasks. Using a 4-billion-parameter LLM backbone, \ours{} achieves state-of-the-art performance on the held-out evaluation benchmark while using only 16 time-series tokens and achieving an average inference latency of 0.15 seconds per question.
\end{abstract}



\received{20 February 2007}
\received[revised]{12 March 2009}
\received[accepted]{5 June 2009}

\maketitle

\section{Introduction}

\begin{figure*}[t]
    \centering
    \begin{subfigure}[t]{0.40\textwidth}
        \raggedright
        \textbf{(a)}\par\vspace{0.2em}
        \centering
        \includegraphics[width=\linewidth]{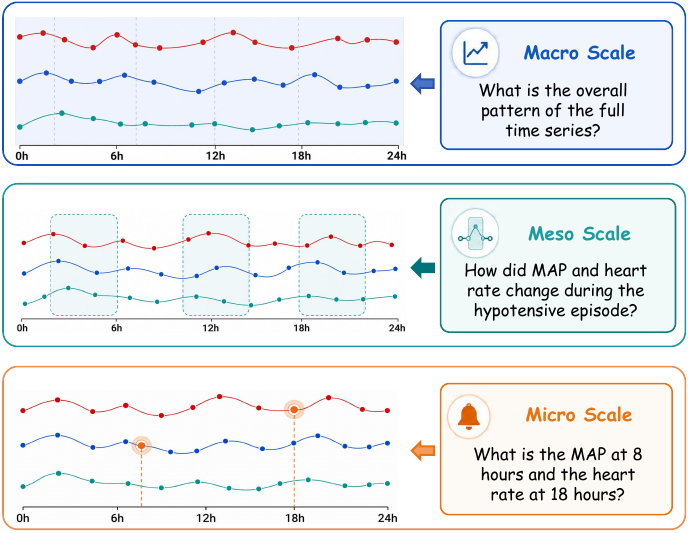}
        \label{fig:intro_scales}
    \end{subfigure}
    \hspace{0.06\textwidth}
    \begin{subfigure}[t]{0.48\textwidth}
        \raggedright
        \textbf{(b)}\par\vspace{0.2em}
        \centering
        \includegraphics[width=\linewidth]{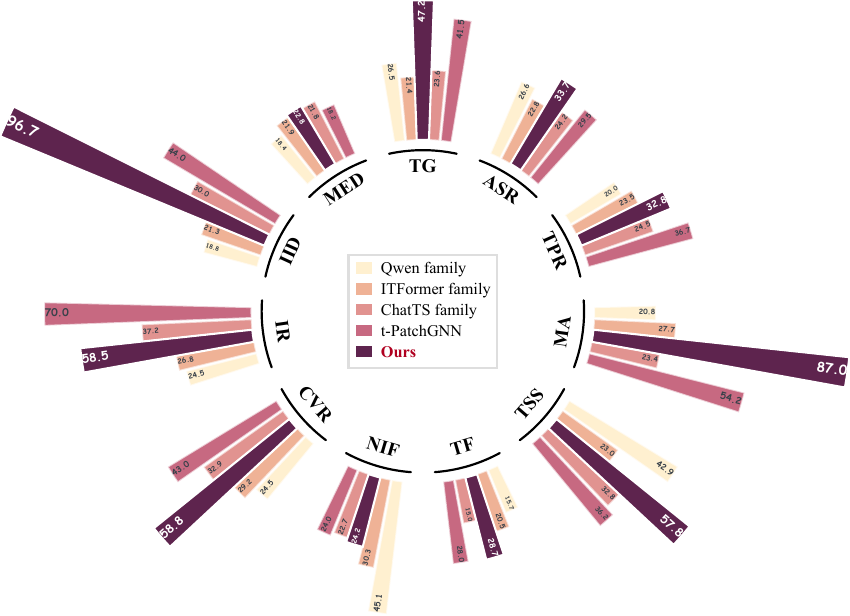}
        \label{fig:intro_results}
    \end{subfigure}
    \caption{Motivation and empirical results for \ours. (a) Clinical questions require evidence at different temporal scales. (b) Existing model families exhibit limited performance across 11 clinical temporal reasoning tasks, suggesting that they fail to consistently capture the sparse and asynchronous nature of irregular clinical time series.}
    \label{fig:intro_motivation}
\end{figure*}

Irregular clinical time series (ICTS) provide detailed records of patient status, disease progression, and treatment responses, making them central to clinical analysis and decision-making~\cite{PhysioNet-mimiciv-3.1,johnson2023mimic}. Existing ICTS models are mainly developed for predefined tasks, such as classification or forecasting~\cite{zhang2023warpformer,zhang2024irregular,liu2025astgi,liu2026rethinking}, which produce outputs in fixed, task-specific formats. In clinical practice, however, users may ask diverse natural-language questions about a patient’s trajectory~\cite{kong2025time,nie2026clirbenchbenchmarkingmultimodalquestion}, such as when deterioration occurred or whether the patient’s condition improved after an intervention. Answering such questions requires models not only to capture temporal patterns but also to identify relevant clinical evidence and express reasoning via language interface.

Recent advances in large language models (LLMs) have demonstrated great potential as a unified interface for temporal data understanding and reasoning~\cite{liang2024foundation,liu2025can,liu2025crossmodalitymodelingtimeseries}. Existing approaches mainly fall into three categories. Text-serialization methods convert numerical sequences into textual prompts that can be directly processed by LLMs~\cite{xue2023promptcast}. However, such methods often require large numbers of tokens and inference costs, and struggle to preserve numerical precision. Representation-adaptation methods overcome this limitation by mapping learned time series representations into the embedding space of LLMs~\cite{jin2024timellmtimeseriesforecasting,liu2024unitime}. For example, Time-LLM~\cite{jin2024timellmtimeseriesforecasting} reprograms time-series patches through textual prototypes to enable forecasting with frozen LLMs. More recent multimodal time-series LLMs, such as Chat-TS~\cite{xie2025chatts} and ITFormer~\cite{wang2025itformer}, further connect temporal encoders with language models via trainable tokens and instruction tuning. Despite fruitful progress, most existing methods focus on regularly sampled sequences and overlook the sparse and asynchronous nature of clinical records.

Specifically, reasoning over irregular clinical trajectories poses three interrelated challenges. (1) \emph{Irregular temporal modeling}: Clinical time series are recorded at nonuniform and variable-specific intervals, and the observation process itself may be informative. Thus, a model needs to jointly capture temporal dynamics and observation patterns. (2) \emph{Sparse multi-scale evidence localization}: Evidence relevant to a clinical question is often sparse and unevenly distributed across a long patient trajectory. Moreover, as illustrated in Figure~\ref{fig:intro_motivation}(a), different questions may depend on evidence at different temporal scales, ranging from global trajectory trends to local event dynamics and individual measurements. The challenge is to identify and preserve complementary evidence across scales while avoiding unnecessary information that increases computational cost. (3) \emph{Irregular temporal-language alignment}: Clinical meaning often emerges from combinations of observations rather than isolated values. However, existing methods provide limited supervision for connecting irregular temporal patterns with language semantics. In addition, large-scale paired datasets containing irregular trajectories and textual descriptions are scarce, making effective temporal-language alignment challenging.

To tackle the above challenges, we propose \ours, a cost-effective multimodal LLM reasoning framework for Question Answering (QA) over ICTS data. Specifically, we first introduce an \emph{irregularity-aware multi-scale encoder}, which effectively captures global trajectory patterns, local event dynamics, and fine-grained observations, respectively. After that, we propose a \emph{temporal evidence distiller} that integrates the outputs at diverse scales and compresses the fused representation into a fixed number of LLM-compatible tokens, enabling efficient interaction between irregular temporal data and language reasoning. Moreover, we devise a \emph{progressive temporal-language alignment} strategy that includes multi-scale encoder pre-training, hierarchical caption alignment, and QA-oriented LLM adaptation.

To support effective training, we construct three complementary resources from de-identified ICU records. First, we curate 30,000 irregular clinical trajectories for multi-scale encoder pre-training. Second, we associate these trajectories with hierarchical descriptions at global, local, and fine-grained levels, providing multi-scale supervision for temporal-language alignment. Third, we construct 41,000 QA instances for QA-oriented instruction tuning, covering 11 tasks across four capabilities. On an independent held-out benchmark for irregular clinical time-series QA, \ours{} achieves state-of-the-art performance while using only a 4-billion-parameter LLM backbone, 16 temporal tokens per question, and an average inference latency of 0.15 seconds.

Our contributions are summarized as follows. (1) We propose an irregularity-aware multi-scale interface that captures sparse temporal evidence and distills variable-length irregular trajectories into a compact sequence of LLM-compatible tokens for efficient reasoning. (2) We develop a progressive training strategy that sequentially aligns irregular temporal representations with the semantic space of LLMs. (3) We construct a suite of training resources, including 30,000 clinical trajectories paired  with hierarchical textual descriptions and 41,000 instruction-tuning instances. (4) We conduct extensive experiments on a held-out irregular clinical time-series question answering benchmark. The results show that our framework is both effective and efficient.

\begin{figure*}[t]
    \centering
    \includegraphics[width=0.92\linewidth]{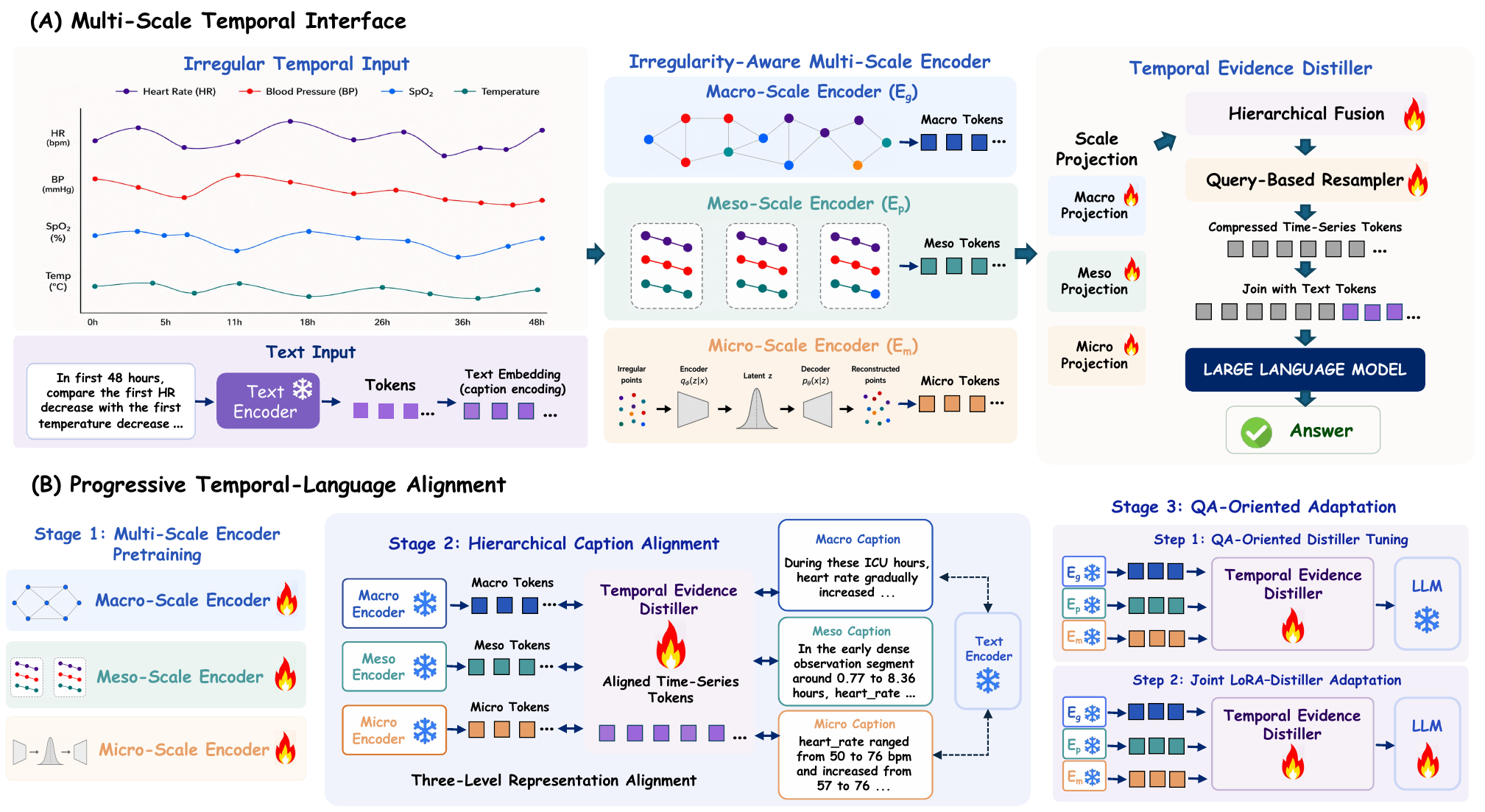}
    \caption{Overview of \ours.
    % Three temporal encoders extract complementary macro, meso, and micro representations directly from irregular observations. The temporal evidence distiller projects and fuses these representations and resamples the valid evidence into a fixed sequence of LLM-compatible tokens. Training proceeds from encoder pretraining to hierarchical caption alignment and QA-oriented adaptation, whereas inference uses only the frozen encoders, the trained distiller, and the adapted LLM.
    }
    \label{fig:framework_overview}
\end{figure*}

\section{Methodology}
\label{sec:method}

\subsection{Problem Formulation}
\label{sec:problem}

An irregular clinical time series can be represented as a collection of univariate sequences, denoted by $\mathcal{X}=\{\mathbf{x}^n\}_{n=1}^N$, where $N$ is the number of clinical variables. Each sequence $\mathbf{x}^n=\{(t_i,x_i)\}_{i=1}^{L_n}$ contains $L_n$ observations of the $n$-th variable, where $t_i$ and $x_i$ are the observed time and value respectively. Building upon this, we define question answering over irregular clinical time series as the task of generating a natural language answer $\mathcal{A}$ conditioned on (1) a natural
language question $q$, (2) a patient-specific irregular clinical time series $\mathcal{X}$, and (3) optional intervention context $\mathbf{c}$, such as treatment and medication. Formally, the task can be expressed as $\mathcal{F}: (q,\mathcal{X},\mathbf{c}) \mapsto \mathcal{A}$, which requires the model $\mathcal{F}(\cdot)$ to interpret the question $q$ and reason over irregular trajectory $\mathcal{X}$ and optional context $\mathbf{c}$ in order to produce the answer $\mathcal{A}$. 

% Let an irregular multivariate time series be denoted as
% \begin{equation}
%     \mathbf{X}=\{(\tau_n,\mathbf{x}_n,\mathbf{m}_n)\}_{n=1}^{N},
% \end{equation}
% where $\tau_n$ is the timestamp, $\mathbf{x}_n\in\mathbb{R}^{D}$ contains the values of $D$ variables, and $\mathbf{m}_n\in\{0,1\}^{D}$ is the observation mask. The timestamps are not assumed to be evenly spaced, and each variable may follow its own missingness pattern.

% Each QA instance contains a question $q$, a candidate answer set $\mathcal{O}=\{o_1,\ldots,o_M\}$, optional auxiliary context $u$, and a gold option index $y\in\{1,\ldots,M\}$. Let $a_y$ denote the target token sequence representing the gold option, such as its option label or answer text. The model estimates
% \begin{equation}
% \small
% p_{\Theta}\left(a_y\mid\mathbf{X},q,\mathcal{O},u\right),
% \end{equation}
% where $\Theta$ denotes the trainable model parameters. This formulation targets a broad range of time-series question answering tasks, including temporal understanding, temporal forecasting, temporal reasoning, and temporal decision-making. Unlike text-only prompting, \ours{} keeps $\mathbf{X}$ in a continuous representation space and communicates temporal information to the LLM through trainable time-series tokens.

\subsection{Framework Overview}
Figure 2 illustrates the overall framework of ClinPRISM, which consists of three major modules. First, an irregularity-aware multi-scale encoder captures sparse temporal evidence at the macro, meso, and micro scales directly from clinical records. Then, a temporal evidence distiller connects the multi-scale encoder and LLM by transforming the learned evidence into a fixed number of temporal tokens. Finally, a progressive temporal–language alignment strategy maps the temporal tokens into the LLM’s semantic space and sequentially adapts them for downstream multi-task reasoning. During inference, the  temporal tokens are inserted into the textual prompt, enabling the LLM to select question-relevant evidence and generate the answer.
% As illustrated in Figure~\ref{fig:framework_overview}, \ours{} is a hierarchical time-series question answering framework for reasoning over irregular multivariate temporal observations with LLMs. It consists of three scale-specialized temporal encoders, a temporal evidence distiller, and an LLM backbone. The encoders extract complementary macro, meso, and micro representations, while the distiller projects, fuses, and compresses them into a fixed-length sequence of LLM-compatible time-series tokens. This compression is question-agnostic and preserves compact evidence about the trajectory; after the temporal tokens are inserted into the prompt, the LLM performs question-conditioned evidence selection and answer generation through attention to the question and candidate options. \ours{} is trained in three stages: temporal encoder pretraining, caption-guided distiller alignment, and downstream QA adaptation through distiller tuning followed by joint LoRA--distiller optimization.

\subsection{Irregularity-Aware Multi-Scale Encoder}
\label{sec:architecture}

Clinical evidence in an irregular trajectory is distributed unevenly across time and variables. A single temporal scope may therefore miss short-lived events or discard the broader context needed to interpret them. We address this limitation by formulating temporal encoding as a scale-conditioned transformation of a common observation set. Each valid measurement is first represented under a shared timestamp-aware event schema; scale-specific aggregation then organizes these events over the complete trajectory at the macro scale, density-adaptive segments at the meso scale, and fine-grained reference times at the micro scale. This design changes how evidence is organized while preserving common observation semantics across scales, providing the temporal evidence distiller with a coherent multi-scale interface. Because every scale operates directly on the irregular sequences defined in Section~\ref{sec:problem}, the encoder preserves the original timestamps and requires neither grid resampling nor explicit imputation.

\textbf{Observation-Centric Interface.} We first convert every valid measurement into a timestamp-aware event representation. For $(t_i,x_i)\in\mathbf{x}^{n}$, we normalize $t_i$ within the trajectory and compute the elapsed time since the previous observation of variable $n$. A scale-specific MLP then encodes the observed value together with a learnable variable embedding, the normalized timestamp, and the observation gap as $\mathbf{z}_{i}^{n,s}$. Since $\mathbf{x}^{n}$ contains only observed measurements, missing or imputed values never enter the encoder. We use $s\in\{g,p,m\}$ to index the macro, meso, and micro scale conditions. Applying the corresponding aggregation at each scale yields a macro vector $\mathbf{h}^{g}\in\mathbb{R}^{d_g}$, a meso sequence $\mathbf{H}^{p}\in\mathbb{R}^{P\times d_p}$, and a micro sequence $\mathbf{H}^{m}\in\mathbb{R}^{R\times d_m}$, where $P$ and $R$ denote the numbers of meso windows and micro reference times.

\textbf{Macro-Scale Encoding.} To retain whole-trajectory context, the macro pathway first applies learnable-query attention pooling within each variable sequence, producing one vector per clinical variable. Multi-head self-attention then models dependencies among variables~\citep{li2025hyperimtshypergraphneuralnetwork}, and a second attention-pooling operation summarizes them as $\mathbf{h}^{g}$. This representation captures long-range trends and cross-variable context that cannot be inferred reliably from an isolated temporal neighborhood.

\textbf{Shared Support-Aware Aggregation.} Localized evidence is not equally reliable across an irregular trajectory: one temporal region may contain many nearby observations, whereas another may be supported by only a few distant measurements. The localized scale conditions therefore use a common aggregation rule that encodes both the local state and its observation support, with scale-specific parameters. For $s\in\{p,m\}$, $\ell_{j,i}^{n,s}$ measures the relevance of the $i$-th observation of variable $n$ to temporal location $j$. We define $w_{j,i}^{n,s}=\exp(\ell_{j,i}^{n,s})$ and $\rho_{j,n}^{s}=\sum_{i=1}^{L_n}w_{j,i}^{n,s}$, where $\rho_{j,n}^{s}$ quantifies the support available for variable $n$ around location $j$. We then compute
\begin{equation}
\small
\begin{aligned}
\mathbf{u}_{j,n}^{s}
&=\operatorname{MLP}_{a}^{s}\!\left(
\frac{\sum_{i=1}^{L_n}w_{j,i}^{n,s}\mathbf{z}_{i}^{n,s}}
{\rho_{j,n}^{s}+\epsilon}
\Vert
\log(1+\rho_{j,n}^{s})
\right),\\
\mathbf{h}_{j}^{s}
&=\operatorname{AttPool}\!\left(
\{\mathbf{u}_{j,n}^{s}:\rho_{j,n}^{s}>0\}
\right).
\end{aligned}
\label{eq:local_encoder}
\end{equation}
The weighted average estimates the local state of variable $n$, while $\log(1+\rho_{j,n}^{s})$ informs the encoder how strongly that estimate is supported by actual observations. Attention pooling then combines the supported variable-level states into $\mathbf{h}_{j}^{s}$. We define $\mu_j^s=\mathbb{I}[\sum_{n=1}^{N}\rho_{j,n}^{s}\geq\eta_s]$, where $\eta_s$ is a branch-specific support threshold, and mask locations with $\mu_j^s=0$ before evidence distillation. The operator thus prevents weakly supported local estimates from being treated as equivalent to densely observed evidence.

\textbf{Meso-Scale Encoding.} To capture episode-level dynamics, the meso pathway instantiates the temporal locations as $P$ learnable soft windows. Each window has a trainable center and positive width and assigns Gaussian relevance according to normalized temporal distance~\citep{liu2026rethinking}. The windows are initialized from an approximately uniform partition and adapt to the empirical observation density during training. Substituting these relevance scores into Equation~\ref{eq:local_encoder} produces $\mathbf{H}^{p}$, whose tokens encode segment-level trends and transitions together with their observation support.

\textbf{Micro-Scale Encoding.} To retain fine-grained temporal states, the micro pathway places $R$ reference times across the normalized observation interval~\citep{shukla2024heteroscedastictemporalvariationalautoencoder}. It computes observation-to-reference relevance through scaled dot-product attention between their time encodings. The resulting scores are substituted into Equation~\ref{eq:local_encoder}, yielding the micro-scale sequence $\mathbf{H}^{m}$. The sequence $\mathbf{H}^{m}$ consequently preserves fine-grained states together with the sampling gaps and observation support around each reference time.

Together, $\mathbf{h}^{g}$, $\mathbf{H}^{p}$, and $\mathbf{H}^{m}$ form a coherent evidence hierarchy for the same irregular trajectory. Their shared event schema and validity semantics allow the temporal evidence distiller to combine broad clinical context, segment-level dynamics, and fine-grained states without reconciling incompatible input representations.

\subsection{Temporal Evidence Distiller}
\label{sec:connector}

The macro, meso, and micro encoders produce representations with heterogeneous structures and dimensions. We therefore introduce a temporal evidence distiller that projects the three scales into a shared LLM space, performs hierarchical cross-scale fusion, and compresses all valid temporal evidence into a fixed-length sequence of temporal tokens.

\textbf{Scale Projection.} We first map the three encoder outputs into the LLM hidden space using scale-specific projection modules:
\begin{equation}
\small
\mathbf{z}^{g}=P_g(\mathbf{h}^{g}),\qquad
\mathbf{Z}^{p}=P_p(\mathbf{H}^{p}),\qquad
\mathbf{Z}^{m}=P_m(\mathbf{H}^{m}).
\label{eq:scale_projection}
\end{equation}
Here, $P_g$, $P_p$, and $P_m$ are trainable projection modules, with $P_p$ and $P_m$ applied independently to each meso and micro token.

\textbf{Hierarchical Fusion.} We use the meso scale as a bridge between macro context and micro temporal evidence. Let $\boldsymbol{\mu}^{p}$ and $\boldsymbol{\mu}^{m}$ denote the validity masks of the meso and micro representations, respectively, and let $\bar{\mathbf{z}}^{m}=\operatorname{MaskedMean}(\mathbf{Z}^{m};\boldsymbol{\mu}^{m})$ denote the pooled micro summary. For the $j$-th meso token, we form the cross-scale context $\mathbf{c}_{j}^{p}=\mathbf{z}^{g}\Vert\mathbf{z}_{j}^{p}\Vert\bar{\mathbf{z}}^{m}$ and compute
\begin{equation}
\small
\tilde{\mathbf{z}}_{j}^{p}
=
\operatorname{LN}\!\left(
\mathbf{z}_{j}^{p}
+
F_{\mathrm{fuse}}(\mathbf{c}_{j}^{p})
\right).
\label{eq:hierarchical_fusion}
\end{equation}
Here, $\operatorname{LN}(\cdot)$ denotes layer normalization, $\Vert$ is feature-wise concatenation, and $F_{\mathrm{fuse}}$ is a lightweight MLP. The residual connection preserves the original meso evidence while incorporating macro and micro context. Applying this operation to all valid meso windows yields the refined sequence $\tilde{\mathbf{Z}}^{p}$.

\textbf{Query-Based Resampling.} Although the numbers of meso and micro positions are fixed, the number of valid tokens varies across trajectories. We concatenate the projected representations along the token dimension as $\mathbf{C}=[\mathbf{z}^{g};\tilde{\mathbf{Z}}^{p};\mathbf{Z}^{m}]_{\mathrm{tok}}$ and construct the corresponding validity mask as $\boldsymbol{\mu}=[1;\boldsymbol{\mu}^{p};\boldsymbol{\mu}^{m}]$. A learned-query resampler then computes
\begin{equation}
\small
\mathbf{T}
=
\operatorname{Resampler}(\mathbf{Q},\mathbf{C};\boldsymbol{\mu})
\in\mathbb{R}^{K\times d_{\mathrm{LLM}}},
\label{eq:temporal_resampling}
\end{equation}
where $\mathbf{Q}\in\mathbb{R}^{K\times d_{\mathrm{LLM}}}$ is a bank of learned, question-independent query tokens. Through masked cross-attention, the resampler extracts information only from valid entries of $\mathbf{C}$ and produces a fixed-length sequence $\mathbf{T}=[\mathbf{t}_{1},\ldots,\mathbf{t}_{K}]$. The resampler is designed to retain compact trajectory-level evidence rather than select evidence for a specific question. Question-conditioned selection occurs subsequently inside the LLM, whose self-attention jointly processes $\mathbf{T}$, the question, and the candidate answers. For fused-caption alignment in Stage 2, we use the mean-pooled representation $\mathbf{z}^{f}=K^{-1}\sum_{k=1}^{K}\mathbf{t}_{k}$.

\textbf{LLM Interface.} The temporal-token sequence $\mathbf{T}$ replaces a reserved time-series placeholder in the LLM input. The remaining prompt contains the question, candidate answers, and optional auxiliary context. In this way, the LLM accesses the irregular trajectory through a compact sequence of continuous temporal embeddings rather than a long textual serialization of the raw observations.

\subsection{Progressive Temporal-Language Alignment}
\label{sec:training}

We train the framework progressively in three stages. Stage 1 learns scale-specialized representations from unlabeled irregular time series. Stage 2 aligns the temporal evidence distiller with hierarchical language supervision while keeping the temporal encoders and the LLM frozen. Stage 3 adapts the aligned temporal tokens to downstream QA through distiller tuning followed by joint LoRA--distiller optimization.

\subsubsection{Stage 1: Multi-Scale Encoder Pretraining}
\label{sec:stage1}

We independently pretrain the macro, meso, and micro branches on the irregular ICU time-series corpus described in Section~\ref{sec:ts_corpus}, using a scale-specific self-supervised objective for each branch. Separate optimization encourages each branch to specialize in its intended temporal scope before cross-scale fusion. No parameters or gradients are shared across branches at this stage, and no caption, QA, or LLM supervision is used. After pretraining, the task-specific heads are removed and the multi-scale encoder remains frozen in all subsequent stages.

\subsubsection{Stage 2: Hierarchical Caption Alignment}
\label{sec:stage2}

The pretrained encoders capture irregular temporal structure, but their outputs are not yet aligned with the contextual language representation space of the LLM. We therefore optimize the temporal evidence distiller using hierarchical captions while keeping both the temporal encoders and the LLM frozen.
For each trajectory, the macro, meso, and micro captions are denoted by $c^{g}$, $c^{p}$, and $c^{m}$, and their concatenation forms the fused caption $c^{f}=\operatorname{concat}(c^{g},c^{p},c^{m})$. We pass each caption through the frozen LLM and mean-pool its final-layer contextual hidden states over non-padding caption tokens, followed by normalization, yielding $\mathbf{e}^{g}$, $\mathbf{e}^{p}$, $\mathbf{e}^{m}$, and $\mathbf{e}^{f}$. On the temporal side, we define $\bar{\mathbf{z}}^{g}=\mathbf{z}^{g}$, obtain $\bar{\mathbf{z}}^{p}$ and $\bar{\mathbf{z}}^{m}$ by masked mean pooling over valid tokens in $\mathbf{Z}^{p}$ and $\mathbf{Z}^{m}$, and set $\bar{\mathbf{z}}^{f}=\mathbf{z}^{f}$.

We use a bidirectional contrastive loss to align each temporal representation with its scale-matched caption representation:
\begin{equation}
\small
\mathcal{L}_{\mathrm{text}}
=
\frac{1}{|\mathcal{S}|}
\sum_{s\in\mathcal{S}}
\mathcal{L}_{\mathrm{NCE}}
\left(
\bar{\mathbf{z}}^{s},
\mathbf{e}^{s}
\right),
\qquad
\mathcal{S}=\{g,p,m,f\},
\label{eq:text_alignment}
\end{equation}
where matched temporal--caption pairs are treated as positives and other pairs in the same mini-batch as negatives.

Because independent caption alignment does not explicitly enforce semantic agreement across temporal scales, we additionally regularize adjacent-scale representations:
\begin{equation}
\small
\mathcal{L}_{\mathrm{scale}}
=
\frac{1}{2}
\left[
1-\cos\left(\bar{\mathbf{z}}^{m},\bar{\mathbf{z}}^{p}\right)
+
1-\cos\left(\bar{\mathbf{z}}^{p},\bar{\mathbf{z}}^{g}\right)
\right].
\label{eq:scale_consistency}
\end{equation}
The Stage-2 objective is $\mathcal{L}_{\mathrm{align}}=\mathcal{L}_{\mathrm{text}}+\lambda_{\mathrm{scale}}\mathcal{L}_{\mathrm{scale}}$. This regularizer acts only on pooled summaries and therefore aligns their coarse semantic direction without forcing individual meso or micro tokens to coincide. Scale-specific caption losses preserve the distinct information associated with each branch, while the frozen, independently pretrained encoders prevent the consistency term from collapsing their underlying temporal representations. Only the temporal evidence distiller is optimized in this stage. The captions serve as training-time semantic supervision and are not used during inference.

\subsubsection{Stage 3: QA-Oriented Adaptation}
\label{sec:stage3}

Stage 2 produces language-aligned temporal tokens, but the framework has not yet been optimized for answering questions. We therefore adapt it to the QA corpus in two steps.

\paragraph{Step 1: QA-Oriented Distiller Tuning.} We first optimize only the temporal evidence distiller while keeping the temporal encoders and the LLM frozen. For QA instances with matched caption supervision, we retain the Stage-2 objectives as auxiliary regularization:
\begin{equation}
\small
\mathcal{L}_{\mathrm{QA}}
=
\mathcal{L}_{\mathrm{ans}}
+
\lambda_{\mathrm{text}}\mathcal{L}_{\mathrm{text}}
+
\lambda_{\mathrm{scale}}\mathcal{L}_{\mathrm{scale}},
\label{eq:stage3_step1}
\end{equation}
where $\mathcal{L}_{\mathrm{ans}}$ is the autoregressive answer language-modeling loss. 

\paragraph{Step 2: Joint LoRA--Distiller Adaptation.} We then introduce LoRA modules into the LLM and jointly optimize the LoRA parameters and the temporal evidence distiller using $\mathcal{L}_{\mathrm{ans}}$. The temporal encoders and the original LLM backbone remain frozen. This step allows the temporal-token interface and the LLM-side reasoning parameters to co-adapt without full LLM fine-tuning.

At inference time, the frozen multi-scale encoder and the trained distiller transform $\mathcal{X}$ into the fixed-length temporal-token sequence $\mathbf{T}$, which replaces the reserved time-series placeholder in the language prompt. The LLM generates $\mathcal{A}$ from the temporal tokens, question $q$, and optional intervention context $\mathbf{c}$; captions and alignment losses are not required.

\begin{table*}[t]
\centering
\setlength{\tabcolsep}{5.0pt}
\renewcommand{\arraystretch}{1.15}
\caption{Model names and values shown in gray denote systems excluded from ranking. Among open-source models with 3--27B parameters, the best and second-best results in each column are shown in bold and underlined, respectively.}
\label{tab:model_task_accuracy}
\scalebox{0.86}{%
\begin{tabular}{lc*{12}{c}}
\toprule
\multicolumn{1}{c}{\multirow{2}{*}{\textbf{Model}}}
& \multirow{2}{*}{\textbf{Params.}}
& \multicolumn{5}{c}{\textbf{Understanding}}
& \multicolumn{2}{c}{\textbf{Forecasting}}
& \multicolumn{2}{c}{\textbf{Reasoning}}
& \multicolumn{2}{c}{\textbf{Decision-Making}}
& \multirow{2}{*}{\textbf{Overall}} \\
\cmidrule(lr){3-7}
\cmidrule(lr){8-9}
\cmidrule(lr){10-11}
\cmidrule(lr){12-13}
&
& \textbf{TG}
& \textbf{ASR}
& \textbf{TPR}
& \textbf{MA}
& \textbf{TSS}
& \textbf{TF}
& \textbf{NIF}
& \textbf{CVR}
& \textbf{IR}
& \textbf{IID}
& \textbf{MED}
& \\
% \midrule
% Human Expert & - & 100.00 & 100.00 & 100.00 & 94.50 & 100.00 & 91.00 & 90.83 & 100.00 & 100.00 & 95.33 & 89.67 & 96.48 \\
\midrule
\rowcolor{tablerowcolor1}
\multicolumn{14}{l}{$\blacktriangledown$ \emph{Closed-source LLMs}} \\
\closed{Gemini-2.5-flash} & -- & \closed{41.67} & \closed{43.33} & \closed{25.00} & \closed{23.33} & \closed{91.67} & \closed{20.00} & \closed{10.00} & \closed{43.33} & \closed{60.00} & \closed{60.00} & \closed{25.00} & \closed{40.30} \\
\closed{GPT-5.4 mini} & -- & \closed{58.33} & \closed{63.33} & \closed{35.00} & \closed{45.00} & \closed{93.33} & \closed{33.33} & \closed{6.67} & \closed{63.33} & \closed{51.67} & \closed{76.67} & \closed{25.00} & \closed{50.15} \\
\midrule
\rowcolor{tablerowcolor}
\multicolumn{14}{l}{$\blacktriangledown$ \emph{Open-source General LLMs}} \\
\closed{DeepSeek-V4-flash} & 284B & \closed{28.33} & \closed{36.67} & \closed{25.00} & \closed{45.00} & \closed{25.00} & \closed{38.33} & \closed{18.33} & \closed{18.33} & \closed{91.67} & \closed{41.67} & \closed{26.67} & \closed{35.91} \\
\closed{KiMi-2.6} & 1T & \closed{50.00} & \closed{56.67} & \closed{26.67} & \closed{73.33} & \closed{58.33} & \closed{36.67} & \closed{16.67} & \closed{15.00} & \closed{93.33} & \closed{91.67} & \closed{21.67} & \closed{49.09} \\
Gemma3-27B-it & 27B & 22.33 & 27.21 & 25.00 & 33.17 & 21.67 & \textbf{31.67} & 20.50 & 16.00 & \textbf{77.83} & 39.67 & \underline{24.50} & 30.87 \\
Gemma4-12B-it & 12B & \underline{45.00} & \textbf{39.50} & 21.17 & 34.33 & 17.67 & 22.17 & 18.50 & 6.17 & 60.33 & 39.67 & 23.50 & 29.82 \\
Qwen3.6-27B~\citep{qwen36plus} & 27B & 23.17 & 27.50 & 12.33 & 14.00 & \textbf{74.00} & 10.50 & 43.33 & 21.67 & 23.83 & 14.33 & 10.50 & 25.01 \\
Qwen3.5-9B~\citep{qwen35blog} & 9B & 27.17 & 25.33 & 26.53 & 23.67 & 30.00 & 18.33 & \underline{45.67} & 27.50 & 27.17 & 20.00 & 21.67 & 26.64 \\
Qwen3.5-4B~\citep{qwen35blog} & 4B & 29.17 & 26.83 & 21.17 & 24.67 & 24.83 & 18.17 & \textbf{46.17} & 24.33 & 22.50 & 22.00 & 23.17 & 25.73 \\
\midrule
\rowcolor{tablerowcolor}
\multicolumn{14}{l}{$\blacktriangledown$ \emph{Time Series Language Models}} \\
TimeOmni-1-7B~\citep{guan2026timeomni1incentivizingcomplexreasoning} & 7B & 23.67 & 29.00 & 16.50 & 12.33 & 14.33 & 14.50 & 15.33 & 10.50 & 11.67 & 25.33 & 7.67 & 16.44 \\
TS-Reasoner-7B~\citep{yu2025tsreasoneraligningtimeseries} & 7B & 29.67 & 28.33 & 27.00 & 28.33 & 26.33 & 27.50 & 23.33 & 27.00 & 38.33 & 35.00 & \textbf{25.00} & 28.71 \\
ITFormer-3B~\citep{wang2025itformer} & 3B & 22.25 & 22.17 & 21.69 & 26.91 & 22.67 & 22.67 & 31.67 & 23.92 & 24.18 & 20.19 & 21.50 & 23.62 \\
ITFormer-7B~\citep{wang2025itformer} & 7B & 20.46 & 23.43 & 25.30 & 28.44 & 23.29 & 18.33 & 28.88 & 34.58 & 29.35 & 22.40 & 22.33 & 25.16 \\
ChatTS-8B~\citep{xie2025chatts} & 8B & 21.83 & 23.00 & 25.33 & 20.17 & 25.83 & 13.00 & 24.67 & 33.83 & 42.67 & 25.67 & 21.50 & 25.23 \\
ChatTS-14B~\citep{xie2025chatts} & 14B & 25.33 & 25.33 & 23.67 & 26.67 & 39.67 & 17.00 & 20.67 & 32.00 & 31.67 & 34.33 & 22.17 & 27.14 \\
Time-LLM~\citep{jin2024timellmtimeseriesforecasting} & 7B & 24.17 & 24.83 & 28.67 & 24.83 & 23.17 & 24.50 & 24.83 & 22.17 & 24.17 & 25.00 & \underline{24.50} & 24.62 \\
AutoTime~\citep{liu2024autotimesautoregressivetimeseries} & 7B & 23.33 & 21.50 & 27.33 & 24.33 & 26.33 & 26.00 & 26.83 & 21.84 & 25.83 & 24.67 & 21.67 & 24.51 \\
t-PatchGNN~\citep{zhang2024irregular}+Qwen3 4B & 4B & 41.50 & 29.50 & \textbf{36.67} & \underline{54.17} & 36.17 & 28.00 & 24.00 & \underline{43.00} & \underline{70.00} & \underline{44.00} & 18.17 & \underline{38.65} \\
\midrule
\rowcolor{blue!5}
\textbf{\ours} & 4B & \textbf{47.17} & \underline{33.67} & \underline{32.83} & \textbf{87.00} & \underline{57.83} & \underline{28.67} & 24.17 & \textbf{58.83} & 58.50 & \textbf{96.67} & 22.83 & \textbf{49.83} \\
\bottomrule
\end{tabular}
}
\end{table*}

\section{Training Corpus Construction}
\label{sec:data_construction}

To support the three-stage training strategy described in Section~\ref{sec:training}, we construct three complementary corpora from the same collection of MIMIC-IV ICU records~\citep{PhysioNet-mimiciv-3.1}: an irregular time-series corpus for temporal encoder pretraining, a hierarchical caption corpus for temporal-language alignment, and a multi-task QA corpus for downstream adaptation. The resulting data consist of 30,000 ICU-stay trajectories, 30,000 matched hierarchical caption records, and approximately 41,000 four-way multiple-choice QA instances.

\subsection{Irregular Clinical Time-Series Corpus}
\label{sec:ts_corpus}

We first construct irregular multivariate time-series instances from MIMIC-IV ICU records, treating each ICU stay as an independent temporal trajectory. When a hospital admission contains multiple ICU stays, each stay is processed separately. Measurements associated with the same stay are mapped to a unified schema of 11 temporal variables, and their timestamps are converted into elapsed hours relative to ICU admission and sorted chronologically. We preserve the original observation times rather than resampling the trajectories onto a regular temporal grid.
For each trajectory, we retain the observed values, variable-level observation masks, and available intervention or event records. This representation preserves nonuniform sampling intervals, variable-specific missingness, and heterogeneous observation frequencies. After preprocessing and leakage control, we select 30,000 ICU-stay trajectories as the shared foundation of all three training corpora. Stage 1 uses their irregular temporal observations for self-supervised pretraining of the three temporal encoders. Variable definitions, preprocessing procedures, benchmark de-overlap checks are provided in Appendix~\ref{apx:ts_construction}.

\subsection{Hierarchical Caption Corpus}
\label{sec:caption_corpus}

We next construct one hierarchical caption record for each of the 30,000 selected trajectories. The construction process first applies deterministic programs to the timestamped observations and observation masks, producing structured summaries at three temporal scales. The global summary describes the complete trajectory, including its duration, observation density, overall trends, and major threshold-defined abnormalities. The segment summary characterizes sufficiently observed local segments distributed across the trajectory. The micro summary records variable-specific behavior, including observation counts, value ranges, start-to-end changes, threshold crossings, and missingness patterns.

GPT-5.5 then verbalizes the three structured summaries into global, segment-level, and micro-level captions. It is explicitly instructed to preserve the supplied variable names, relative timestamps, numerical values, temporal trends, and abnormality indicators, while avoiding diagnoses, causal interpretations, and unsupported clinical information. The fused caption is formed by concatenating the three generated captions rather than by an additional generation step. Finally, the generated records undergo automatic checks for structural completeness and consistency with the underlying summaries. During Stage 2, the global, segment-level, micro-level, and fused captions provide scale-matched semantic supervision for their corresponding temporal representations. The captions are used only as training-time semantic anchors and are not included in the model input at inference time. Further details on structured summarization, generation constraints, and quality control are provided in Appendix~\ref{apx:caption_construction}.

\subsection{Multi-Task Time-Series QA Corpus}
\label{sec:qa_corpus}

We further construct approximately 41,000 four-way multiple-choice QA instances from the selected ICU trajectories. A single trajectory may produce multiple instances when it satisfies the requirements of different task schemas. The corpus covers four complementary capabilities---temporal understanding, temporal forecasting, temporal reasoning, and temporal decision-making---which are further divided into 11 task types. For each task type, we first define an executable schema specifying the queried variables, model-visible time range, reasoning objective, evidence rule, and answer format. Deterministic programs then instantiate the schema over eligible trajectories and compute the temporal anchors, evidence windows, gold answers, and initial question--option structures directly from the timestamped observations or intervention records.

After the temporal evidence and gold labels have been fixed, GPT-5.5 is used only to paraphrase the questions and answer options, increase linguistic diversity, reduce superficial lexical cues, and strengthen the distractors. It is constrained to preserve the original reasoning objective, temporal scope, option meanings, supporting evidence, and gold answer. The rewritten instances subsequently undergo automatic consistency and leakage checks, followed by human verification of task suitability, linguistic quality, evidence support, and answer-label consistency. Instances with unresolved ambiguity or insufficient evidence are removed. The validated QA instances are used for QA-oriented distiller tuning and joint LoRA-distiller adaptation in Stage 3. The complete task taxonomy and validation protocol are provided in Appendix~\ref{apx:qa_construction}.

\begin{table*}[t]
\centering
\caption{Ablation study of different model variants. The best result in each column is shown in bold.}
\label{tab:variant_ablation}
\setlength{\tabcolsep}{5.0pt}
\renewcommand{\arraystretch}{1.15}

\scalebox{0.8}{%
\begin{tabular}{l*{12}{c}}
\toprule
\multicolumn{1}{c}{\multirow{2}{*}{\textbf{Variant}}}
& \multicolumn{5}{c}{\textbf{Understanding}}
& \multicolumn{2}{c}{\textbf{Forecasting}}
& \multicolumn{2}{c}{\textbf{Reasoning}}
& \multicolumn{2}{c}{\textbf{Decision-Making}}
& \multirow{2}{*}{\textbf{Overall}} \\
\cmidrule(lr){2-6}
\cmidrule(lr){7-8}
\cmidrule(lr){9-10}
\cmidrule(lr){11-12}
& \textbf{TG}
& \textbf{ASR}
& \textbf{TPR}
& \textbf{MA}
& \textbf{TSS}
& \textbf{TF}
& \textbf{NIF}
& \textbf{CVR}
& \textbf{IR}
& \textbf{IID}
& \textbf{MED}
& \\
\midrule
Only Macro & 46.00 & 32.17 & 37.00 & 78.83 & 54.50 & 23.33 & 20.83 & 45.67 & 56.50 & \textbf{98.17} & 23.67 & 46.97 \\
Macro + Meso & \textbf{49.17} & 32.17 & 33.50 & 81.83 & 55.83 & 25.67 & 24.00 & 47.00 & 62.67 & 95.50 & 23.33 & 48.24 \\
Without Stage 2 & 47.33 & 31.33 & 33.67 & 83.33 & 55.67 & 27.00 & 22.67 & 56.17 & 53.17 & 98.00 & 22.67 & 48.27 \\
Without Stage 3 Step 1 & 46.33 & 30.67 & \textbf{37.50} & 81.50 & 51.17 & 26.83 & 21.50 & 48.50 & 59.83 & 97.33 & 21.83 & 47.55 \\
Without Stage 3 Step 2 & 35.33 & 26.17 & 31.00 & 24.33 & 32.00 & 26.50 & 23.33 & 47.83 & \textbf{68.33} & 60.00 & \textbf{25.00} & 36.35 \\
Stage 3 Step 1 with $\mathcal{L}_{\mathrm{ans}}$ Only & 44.50 & 32.50 & 35.00 & \textbf{87.00} & 47.00 & 26.83 & 23.00 & 50.67 & 63.33 & 96.33 & 24.00 & 48.20 \\
\midrule
\rowcolor{blue!5}
\textbf{\ours} & 47.17 & \textbf{33.67} & 32.83 & \textbf{87.00} & \textbf{57.83} & \textbf{28.67} & \textbf{24.17} & \textbf{58.83} & 58.50 & 96.67 & 22.83 & \textbf{49.83} \\
\bottomrule
\end{tabular}%
}
\end{table*}

\section{Experiment}
\subsection{Experimental Setup}
\textbf{Datasets and metrics.} We train \ours{} on the QA corpus constructed in Section~\ref{sec:data_construction} and evaluate it on CLIR-Bench~\citep{nie2026clirbenchbenchmarkingmultimodalquestion}, a held-out benchmark for irregular multivariate time-series QA. CLIR-Bench contains 11 tasks grouped into four capability dimensions: temporal understanding, including Temporal Grounding (TG), Anchor-State Retrieval (ASR), Trend Pattern Recognition (TPR), Missingness Awareness (MA), and Clinical Time-Series Summarization (TSS); temporal reasoning, including Asynchronous Cross-Variable Reasoning (CVR) and Intervention Response (IR); temporal forecasting, including Threshold Forecasting (TF) and Next-Value Interval Forecasting (NIF); and temporal decision-making, including Immediate Intervention Decision (IID) and Monitoring/Escalation Decision (MED). Since all questions follow a multiple-choice format, we report accuracy as the evaluation metric.

\textbf{Baselines.} We compare \ours{} with a diverse set of baseline models, including (1) closed-source LLMs: Gemini-2.5-flash and GPT-5.4 mini; (2) open-source general LLMs: DeepSeek-V4-flash, KiMi-2.6, Gemma3-27B-it, Gemma4-12B-it, Qwen3.5-4B~\citep{qwen35blog}, Qwen3.5-9B~\citep{qwen35blog}, and Qwen3.6-27B~\citep{qwen36plus}; (3) time-series LLMs: TimeOmni-1~\citep{guan2026timeomni1incentivizingcomplexreasoning}, TS-Reasoner~\citep{yu2025tsreasoneraligningtimeseries}, ITFormer~\citep{wang2025itformer}, ChatTS~\citep{xie2025chatts}, AutoTime~\citep{liu2024autotimesautoregressivetimeseries}, and TimeLLM~\citep{jin2024timellmtimeseriesforecasting}; and (4) an irregular time series LLM using t-PatchGNN~\citep{zhang2024irregular} as the time-series encoder. For AutoTime and TimeLLM, we follow the ITFormer~\citep{wang2025itformer} adaptation setting, so that they can process time-series inputs under a unified configuration. In addition, the t-PatchGNN baseline is implemented on Qwen3-4B~\citep{yang2025qwen3technicalreport} and follows the same training pipeline as \ours. The only difference is that this baseline uses t-PatchGNN as the time-series encoder without caption alignment.

\textbf{Training Details.} We train the full framework in three stages as shown in Figure~\ref{fig:framework_overview}. In Stage 1, the temporal encoders are pretrained on the irregular time-series corpus constructed in Section~\ref{sec:ts_corpus}. In Stage 2, the pretrained encoders are frozen, and the hierarchical caption data constructed in Section~\ref{sec:caption_corpus} are used to train the time-series connector for temporal-textual alignment. The connector outputs a fixed-length sequence \(T\) of LLM-compatible time-series tokens, with \(|T|=16\), which are inserted into the LLM input. In Stage 3, we use the multi-task QA corpus, which contains approximately 41,000 multiple-choice questions grounded in irregular multivariate time-series records, to tune the connector and adapt the language model with LoRA, respectively. Both Stage 3 steps are trained for 2 epochs,  $\lambda_{\mathrm{scale}}=0.05$ in Stage 2, $\lambda_{\mathrm{text}}=0.1$ and $\lambda_{\mathrm{scale}}=0.01$ in Stage 3 Step 1. All language-model stages are built on Qwen3-4B~\citep{yang2025qwen3technicalreport}.

\begin{figure*}[t]
\centering
\includegraphics[width=\textwidth]{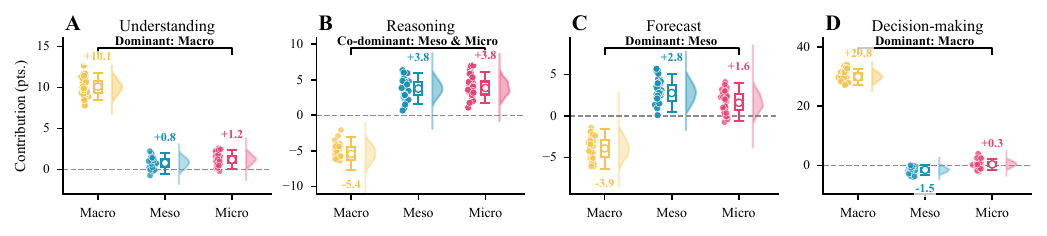}
\caption{Capability-level reliance on hierarchical temporal representations. Panels show bootstrapped accuracy-point contributions from Macro, Meso, and Micro representations. Points denote bootstrap samples, boxes summarize quartiles, and half-violins show density. Panel-specific vertical scales emphasize the strongest scale, with ties reported jointly.}
\label{fig:temporal_layer_reliance}
\end{figure*}

\subsection{Main Results}
Table~\ref{tab:model_task_accuracy} reports the results on CLIR-Bench. \ours{} achieves the highest macro-average accuracy among all open-source systems, reaching 49.83. It outperforms the strongest time-series-oriented baseline, t-PatchGNN, by 11.18 points and the best open-source general LLM, KiMi-2.6, by 0.74 points. Despite using only a 4B LLM backbone, \ours{} trails GPT-5.4 mini by only 0.32 percentage points and exceeds Gemini-2.5-flash by 9.53 points. The gains are especially clear on MA, CVR and IID, suggesting that hierarchical temporal representations and caption-guided alignment help the model capture sparse observations, multi-scale trends, and clinically relevant temporal evidence.

\begin{table}[t]
\centering
\caption{Sensitivity to temporal-token budget $K$.}
\label{tab:token_budget_paper}
\setlength{\tabcolsep}{3.0pt}
\renewcommand{\arraystretch}{1}

\begin{tabular}{cccccc}
\toprule
\textbf{$K$} & 8 & \textbf{16} & 32 & 64 & 128 \\
\midrule
\textbf{Average} & 46.02 & \textbf{49.83} & 47.79 & 47.59 & 47.45 \\
\bottomrule
\end{tabular}
\end{table}

\subsection{Ablation Studies}

Table~\ref{tab:variant_ablation} reports the ablation results for the multi-scale encoders and progressive training strategy. The macro-only, macro-plus-meso, and full models achieve average accuracies of 46.97, 48.24, and 49.83, respectively, showing that meso- and micro-scale representations provide complementary temporal evidence. Removing Stage~2, Stage~3 Step~1, or Stage~3 Step~2 reduces the average accuracy to 48.27, 47.55, and 36.35, respectively, with the largest drop occurring without joint LoRA--connector adaptation. Using only $\mathcal{L}_{\mathrm{ans}}$ in Stage~3 Step~1 yields 48.20, indicating that the auxiliary alignment objectives remain beneficial during QA adaptation. As shown in Table~\ref{tab:token_budget_paper}, we further analyze the sensitivity to the temporal-token sequence length $K=|\mathbf{T}|$. The best overall performance is achieved with $K=16$, and additional task-level results are in Appendix~\ref{apx:token_budget}.

Figure~\ref{fig:temporal_layer_reliance} decomposes capability-level gains across successive encoder variants. Macro measures the improvement over t-PatchGNN, whereas Meso and Micro measure the gains from adding the corresponding encoders. Macro contributes most to understanding and decision-making ($+10.1$ and $+29.8$ points), indicating the importance of whole-record context. Meso contributes most to forecasting ($+2.8$), while Meso and Micro contribute nearly equally to reasoning ($+3.8$ each). These patterns show that broad context and localized dynamics provide complementary evidence across QA capabilities.

\begin{figure}[t]
    \centering
    \includegraphics[width=0.85\linewidth]{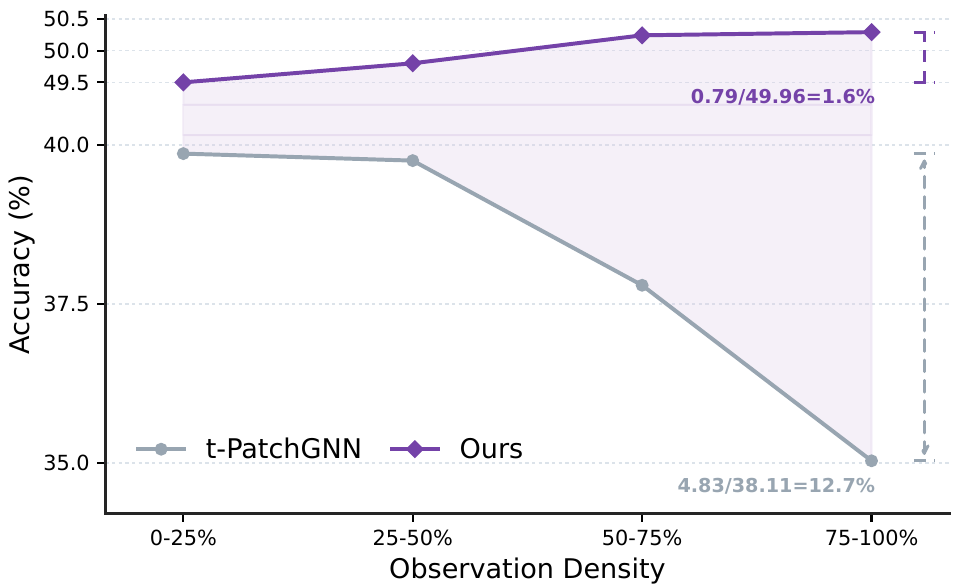}
    \caption{Accuracy under different observation-density ranges. \ours{} maintains robust performance across observation-density levels.}
    \label{fig:density_robustness}
\end{figure}

\subsection{Robustness to Observation Density}
To evaluate robustness under varying sparsity, we group CLIR-Bench samples by observation density, defined as the proportion of observed variable--time cells in each trajectory. After normalizing the observed density range, we divide it into four equal-width intervals and report QA accuracy within each interval. We quantify stability as $(A_{\max}-A_{\min})/\bar{A}$, where $A_{\max}$, $A_{\min}$, and $\bar{A}$ denote the maximum, minimum, and mean interval accuracies, respectively; lower values indicate less sensitivity to observation density.
As shown in Figure~\ref{fig:density_robustness}, \ours{} outperforms t-PatchGNN in all four intervals and exhibits substantially lower relative variation, with $1.6\%$ compared with $12.7\%$. These results indicate that \ours{} maintains stable QA performance across trajectories with different observation densities, including highly sparse inputs.

\begin{figure}[t]
    \centering
    \includegraphics[width=0.88\linewidth]{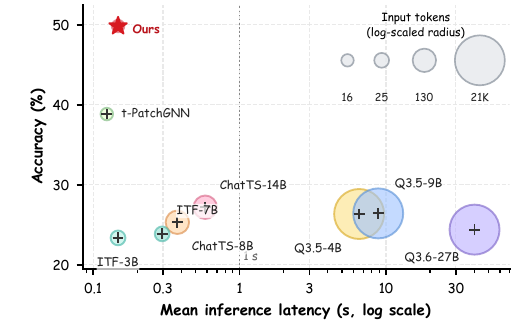}
    \caption{Accuracy-efficiency trade-off. 
    % The horizontal axis shows QA accuracy, the vertical axis shows mean per-question inference latency on a logarithmic scale, and marker size represents input temporal-token cost.
    }
    \label{fig:input_efficiency_tradeoff}
\end{figure}

\subsection{Accuracy--Efficiency Trade-off}
Figure~\ref{fig:input_efficiency_tradeoff} compares QA accuracy, mean per-question inference latency, and token consumption. \ours{} achieves $49.83\%$ accuracy with a latency of $0.15$ seconds and only 16 temporal tokens per question. Under the same 16-token budget, t-PatchGNN is $0.03$ seconds faster but achieves $11.18$ percentage points lower accuracy.
\ours{} also substantially outperforms ITFormer and ChatTS while using comparable or lower inference latency and fewer temporal tokens. The text-serialized Qwen baselines require approximately 21K input tokens and $6.6$--$40$ seconds per question, yet achieve only $25.01\%$--$26.64\%$ accuracy. In comparison, \ours{} reaches $49.83\%$ accuracy with only 16 temporal tokens and $0.15$ seconds of inference latency. Overall, \ours{} achieves the highest QA accuracy while retaining low inference latency and a compact temporal-token interface. Detailed configurations are provided in the Appendix~\ref{apx:inference latency}.

\section{Related Work}
\label{sec:related_work}

\subsection{Time-Series Question Answering}
\label{sec:rw_tsqa}

Time-Series Question Answering~(TSQA) has recently shown great promise for querying temporal data with LLMs~\citep{divo2025quantsquestionansweringtime, zhang2024largelanguagemodelstime,liu2025crossmodalitymodelingtimeseries}. Recent work has begun to formulate TSQA as a unified temporal-textual learning problem. Time-MQA~\citep{kong2025time} builds a large multi-task TSQA corpus covering forecasting, imputation, anomaly detection, classification, and open-ended reasoning across multiple domains. Chat-TS~\citep{xie2025chatts} introduces time-series tokens and instruction data to support multi-task QA over time-series and textual inputs. MTBench~\citep{chen2026mtbenchmultimodaltimeseries} evaluates temporal reasoning and question answering over paired time-series and textual context in finance and weather domains. ChatTime~\citep{wang2024chattimeunifiedmultimodaltime} studies multi-task time-series QA over synthetic temporal patterns. 
EngineMT-QA~\citep{wang2025itformer} focuses on aero-engine signal question answering for temporal understanding, reasoning, and decision-making. TSAQA~\citep{jing2026tsaqatimeseriesanalysis} broadens TSQA evaluation with diverse analysis tasks such as characterization, comparison, transformation, and temporal relationship analysis. ARFBench~\citep{xie2026arfbenchbenchmarkingtimeseries} evaluates anomaly-oriented TSQA for software incident response using production telemetry. Despite recent progress, most TSQA settings assume regularly sampled inputs, leaving question answering over irregular clinical trajectories underexplored.

\subsection{Time-Series Multimodal LLMs}
\label{sec:rw_ts_mllm}

Time-series MLLMs bridge temporal signals and LLMs beyond raw-text serialization. 
% PromptCast~\citep{xue2023promptcast} convert numerical sequences into textual prompts or digit strings for LLM forecasting. 
GPT4TS~\citep{zhou2023fitsallpowergeneraltime} adapts pretrained language models to general time-series analysis by reusing frozen backbone layers. UniTime~\citep{liu2024unitime} uses domain instructions and a Language-TS Transformer for cross-domain time-series representation learning. TEMPO~\citep{cao2024tempopromptbasedgenerativepretrained} incorporates textual information or prompt-based adaptation for multimodal time-series forecasting. ITFormer~\citep{wang2025itformer} introduces learnable instruction tokens and temporal-channel attention to align time-series representations with frozen LLMs. However, most existing time-series MLLMs are designed for regularly sampled inputs or fixed prediction tasks. In this paper, we propose ClinPRISM, a multimodal LLM reasoning framework for irregular time-series QA that combines multi-scale temporal modeling with progressive temporal-language alignment.

\section{Conclusion}

In this work, we proposed \ours, a cost-effective multimodal LLM framework for question answering over irregular clinical time series. By combining global, event-level, and point-level temporal representations, \ours{} captures complementary evidence from sparse and asynchronous observations and distills it into a small set of LLM-compatible tokens. A progressive alignment strategy further connects temporal representations with language semantics and downstream reasoning tasks. Experiments show that \ours{} achieves strong accuracy and an effective accuracy–efficiency trade-off using only a 4B LLM, 16 temporal tokens, and 0.15 seconds of inference time per question. Future work will extend the framework to open-ended questions, uncertainty-aware reasoning, and evaluation across broader clinical settings.

\bibliographystyle{ACM-Reference-Format}
\bibliography{sample-base}

\clearpage
\appendix

\section{Ethics and Privacy Statement}
This study uses de-identified MIMIC-IV ICU records and involves no direct interaction with patients. All data were accessed and processed in accordance with the applicable MIMIC-IV and PhysioNet data-use requirements. Patient, admission, and ICU-stay identifiers were excluded from the inputs provided to language models. Only task-relevant temporal records, timestamps, and intervention information were used. The generated captions were automatically checked, while QA instances were further verified by human reviewers. ClinPRISM is intended solely for research and is not designed for autonomous clinical decision-making. Any real-world deployment would require independent external validation, privacy and fairness assessment, and continuous oversight by qualified healthcare professionals.

% This study uses de-identified MIMIC-IV ICU records and involves no direct interaction with patients. Patient, admission, and ICU-stay identifiers were excluded from GPT-5.5 inputs; generated captions were automatically checked, while QA instances were further verified by human reviewers. ClinPRISM is intended solely for research rather than autonomous clinical use, and any real-world deployment would require external validation, safety and fairness assessment, and clinician oversight.

\section{Details of Training Corpus Construction}
\label{apx:training_data_construction}

This appendix provides the implementation details of the three corpus-construction procedures summarized in Section~\ref{sec:data_construction}. We describe the preprocessing and leakage control used for the irregular ICU time-series corpus, the structured summarization and constrained verbalization used for the hierarchical caption corpus, and the rule-grounded generation and validation procedures used for the multi-task QA corpus.

\subsection{Construction Details of the Irregular ICU Time-Series Corpus}
\label{apx:ts_construction}

We construct the irregular time-series corpus from MIMIC-IV ICU records~\citep{PhysioNet-mimiciv-3.1}. Each instance corresponds to a single ICU stay rather than an entire hospital admission. Measurements and events associated with the same ICU stay are collected from the source tables and mapped to a unified variable schema. Their timestamps are converted into elapsed hours relative to ICU admission and sorted chronologically. We preserve the original observation times and do not resample the trajectories onto a regular temporal grid.

The corpus contains the 11 temporal variables summarized in Table~\ref{tab:ts_variables}.
Each processed trajectory contains an ordered sequence of ICU-relative timestamps, the values of the 11 variables at those timestamps, and a corresponding observation mask for each variable. At a given timestamp, the mask is set to one when the variable is observed and zero otherwise. Unobserved entries are stored as missing values and are distinguished from valid numerical measurements through the masks. For each variable, we additionally record the elapsed time since its previous valid observation. When available, intervention and event records occurring during the ICU stay are retained as auxiliary information.

After timestamp conversion, chronological ordering, and mask construction, we select 30,000 ICU-stay trajectories to form the final corpus. The resulting trajectories preserve the nonuniform sampling intervals, variable-specific missingness, and heterogeneous observation frequencies of the original ICU records. The data partitions are fixed before any captions or QA instances are constructed, and all derived records inherit the partition of their source trajectory.

Each trajectory retains valid \texttt{subject\_id}, \texttt{hadm\_id}, and \texttt{stay\_id} identifiers for leakage control. To ensure separation from CLIR-Bench, we compare the selected corpus with the benchmark using all three identifiers. None of the selected trajectories shares a \texttt{subject\_id}, \texttt{hadm\_id}, or \texttt{stay\_id} with any CLIR-Bench trajectory. Therefore, the training corpus and CLIR-Bench contain no shared patients, hospital admissions, or ICU stays.

\begin{table}[t]
\centering
\small
\caption{Clinical variables used in the irregular ICU time-series corpus.}
\label{tab:ts_variables}
\begin{tabular}{cc}
\toprule
\textbf{Variable} & \textbf{Meaning} \\
\midrule
$\mathrm{heart\_rate}$ & Heart rate \\
$\mathrm{map}$ & Mean arterial pressure \\
$\mathrm{systolic\_bp}$ & Systolic blood pressure \\
$\mathrm{diastolic\_bp}$ & Diastolic blood pressure \\
$\mathrm{spo2}$ & Peripheral oxygen saturation \\
$\mathrm{resp\_rate}$ & Respiratory rate \\
$\mathrm{temperature}$ & Body temperature \\
$\mathrm{lactate}$ & Blood lactate level \\
$\mathrm{creatinine}$ & Serum creatinine level \\
$\mathrm{wbc}$ & White blood cell count \\
$\mathrm{urine\_output}$ & Urine output \\
\bottomrule
\end{tabular}
\end{table}

\subsection{Construction Details of the Hierarchical Caption Corpus}
\label{apx:caption_construction}

We construct one hierarchical caption record for each of the 30,000 selected ICU trajectories. The construction procedure consists of deterministic temporal summarization, constrained natural-language verbalization, and automatic quality control.

For each trajectory, we first compute structured summaries directly from its timestamps, observed values, and observation masks. Missing entries are excluded from all numerical calculations. The summaries are organized at three temporal scales. The global summary records the start and end times, total duration, overall observation density, sequence-level trends, and threshold-defined abnormalities of the complete trajectory. The patch summary describes sufficiently observed local segments distributed chronologically across the trajectory, including their temporal ranges, observed variables, dominant changes, and threshold events. The micro summary describes each variable separately, including its valid observation count, value range, first and last observed values, overall direction of change, threshold crossings, and missingness pattern.

Only information recoverable from valid observations is included in the structured summaries. Variables or local segments with insufficient measurements are marked as unavailable and are not assigned a temporal trend. Patient, hospital-admission, and ICU-stay identifiers are excluded from the generation input. GPT-5.5 receives only an anonymous episode identifier and the three structured temporal summaries.

GPT-5.5 then converts the global, segment-level, and micro-level summaries into their corresponding natural-language captions. The generation prompt requires it to preserve all supplied variable names, relative timestamps, numerical values, temporal trends, and abnormality indicators. It is explicitly instructed not to introduce diagnoses, causal interpretations, unobserved events, or other unsupported clinical information. The generated output contains three fields: \texttt{global\_caption}, \texttt{patch\_caption}, and \texttt{micro\_caption}. The fused caption is subsequently constructed by concatenating the validated global, segment-level, and micro-level captions; it is not generated independently by GPT-5.5.

Each generated record undergoes automatic structural and factual consistency checks. We verify that the record contains a valid anonymous episode identifier, all required output fields, nonempty captions, and only variables present in the structured input. Numerical values, temporal ranges, trend descriptions, and abnormality indicators appearing in the captions are compared with the corresponding structured summaries. Outputs containing unsupported variables, altered numerical information, inconsistent timestamps, malformed fields, or additional clinical conclusions are rejected and regenerated.
After validation, the hierarchical caption corpus contains 30,000 records paired one-to-one with their source trajectories. Each record contains global, segment-level, micro-level, and fused descriptions of the same irregular ICU trajectory.

In addition to automatic validation, we sampled 3,000 hierarchical-caption records, accounting for 10\% of the caption corpus, for human verification. Three master’s students specializing in artificial intelligence and with strong English proficiency independently reviewed the sampled records for numerical and temporal consistency with the source observations, the absence of unsupported clinical information, and consistency across the global-, segment-, and micro-level captions. Any disagreements were resolved through discussion until consensus was reached.

\subsection{Construction Details of the Multi-Task Time-Series QA Corpus}
\label{apx:qa_construction}

We construct approximately 41,000 four-way multiple-choice QA instances from the selected ICU trajectories. A single trajectory may produce multiple QA instances when it satisfies the eligibility requirements of different task schemas. Each instance contains a time-series identifier, a natural-language question, four candidate answers, one gold option, and task-specific temporal anchors or evidence metadata.

The QA corpus covers four complementary capabilities and 11 task types, as summarized in Table~\ref{tab:qa_tasks}. Temporal understanding evaluates the localization, retrieval, recognition, and summarization of observed temporal states. Temporal forecasting requires predicting future measurements from observations available before a predefined cutoff. Temporal reasoning evaluates relationships across variables and intervention events. Temporal decision-making associates recent temporal evidence with intervention or monitoring choices.

\begin{table}[t]
\centering
\caption{Overview of the 11 task types in the multi-task time-series QA corpus.}
\label{tab:qa_tasks}
\small
\setlength{\tabcolsep}{5pt}
\renewcommand{\arraystretch}{1.15}
\scalebox{0.85}{%
\begin{tabular}{
    >{\centering\arraybackslash}p{3.2cm}
    >{\centering\arraybackslash}p{4.5cm}
}
\toprule
\textbf{Capability} & \textbf{Task} \\
\midrule
\multirow[c]{5}{3.2cm}{\centering Temporal Understanding}
& Temporal Grounding \\
& Anchor-State Retrieval \\
& Trend Pattern Recognition \\
& Missingness Awareness \\
& Clinical Time-Series Summarization \\
\midrule
\multirow[c]{2}{3.2cm}{\centering Temporal Forecasting}
& Threshold Forecasting \\
& Next-Value Interval Forecasting \\
\midrule
\multirow[c]{2}{3.2cm}{\centering Temporal Reasoning}
& Cross-Variable Reasoning \\
& Intervention Response \\
\midrule
\multirow[c]{2}{3.2cm}{\centering Temporal Decision-Making}
& Immediate Intervention Decision \\
& Monitoring / Escalation Decision \\
\bottomrule
\end{tabular}
}
\end{table}

For each task type, we first define an executable schema specifying the trajectory eligibility criteria, queried variables, model-visible time range, reasoning objective, evidence rule, and answer format. Deterministic programs instantiate these schemas over eligible trajectories and identify the temporal anchors and evidence windows required by the task. The gold answer is then computed directly from the corresponding timestamped observations or intervention records. The initial question and candidate options are generated only after the evidence window and gold label have been fixed.

Distractors are selected from plausible alternatives within the same task family, including nearby value intervals, adjacent temporal windows, confusable trend patterns, and plausible but unsupported intervention choices. Because the evidence and gold answer are determined programmatically before linguistic rewriting, GPT-5.5 does not independently infer or modify the answer label.

For forecasting tasks, each instance contains a cutoff time that strictly separates the model-visible history from the future label window. Only observations occurring before the cutoff are included in the model input, whereas the gold answer is computed exclusively from valid observations in the future window. Trajectories without sufficient pre-cutoff context or valid post-cutoff evidence are excluded.

After the initial questions, options, evidence metadata, and gold labels have been fixed, GPT-5.5 performs meaning-preserving rewriting. It paraphrases the questions and candidate answers, increases linguistic diversity, reduces superficial lexical cues, and improves the plausibility of distractors. The rewriting prompt explicitly requires preservation of the original task type, reasoning objective, temporal scope, option meanings, supporting evidence, and gold answer. GPT-5.5 therefore modifies only the linguistic expression of an instance and does not generate or revise its underlying evidence or label.

The rewritten instances next undergo automatic consistency checks. We remove samples containing inconsistent timestamps, invalid evidence windows, insufficient observations, missing intervention context, duplicated or malformed options, multiple valid answers, or evidence that cannot be recovered from the source trajectory. We also verify that every instance contains exactly four mutually exclusive options and one valid gold label. For forecasting tasks, an additional leakage check ensures that post-cutoff observations are absent from the model-visible input. The answer options are then randomly shuffled to reduce position bias, and the gold-option index is updated accordingly.

\begin{table*}[t]
\centering
\caption{Sensitivity to the temporal-token budget $K$ on CLIR-Bench. Here, $K=|\mathbf{T}|$ denotes the number of tokens in the resampled temporal-token sequence $\mathbf{T}$. Avg. denotes the macro-average accuracy across all 11 tasks, and bold indicates the best result in each column.}
\label{tab:token_budget}
\setlength{\tabcolsep}{5.0pt}
\renewcommand{\arraystretch}{1.15}

\scalebox{0.90}{%
\begin{tabular}{c*{12}{c}}
\toprule
\textbf{$T$}
& \multicolumn{5}{c}{\textbf{Understanding}}
& \multicolumn{2}{c}{\textbf{Forecasting}}
& \multicolumn{2}{c}{\textbf{Reasoning}}
& \multicolumn{2}{c}{\textbf{Decision-Making}}
& \textbf{Overall} \\
\cmidrule(lr){2-6}\cmidrule(lr){7-8}\cmidrule(lr){9-10}\cmidrule(lr){11-12}\cmidrule(lr){13-13}
& \textbf{TG} & \textbf{ASR} & \textbf{TPR} & \textbf{MA} & \textbf{TSS}
& \textbf{TF} & \textbf{NIF}
& \textbf{CVR} & \textbf{IR}
& \textbf{IID} & \textbf{MED}
& \textbf{Avg.} \\
\midrule
8 & \textbf{48.83} & 31.50 & 33.00 & 70.33 & 52.50 & 26.17 & 24.50 & 48.83 & 52.00 & \textbf{97.83} & 20.67 & 46.02 \\
\textbf{16} & 47.17 & \textbf{33.67} & 32.83 & \textbf{87.00} & \textbf{57.83} & \textbf{28.67} & 24.17 & \textbf{58.83} & 58.50 & 96.67 & 22.83 & \textbf{49.83} \\
32 & 46.00 & 31.50 & 33.67 & 84.33 & 46.67 & 28.17 & 23.67 & 54.67 & 53.67 & 97.17 & 26.17 & 47.79 \\
64 & 46.00 & 32.50 & \textbf{37.33} & 75.50 & 42.33 & 28.17 & 24.17 & 51.00 & \textbf{61.00} & 96.00 & \textbf{29.50} & 47.59 \\
128 & 47.00 & 30.67 & 31.00 & 84.67 & 46.83 & 26.83 & \textbf{26.00} & 53.00 & 56.67 & 95.83 & 23.50 & 47.45 \\
\bottomrule
\end{tabular}%
}
\end{table*}

Finally, the remaining instances are verified by three master's students specializing in artificial intelligence and possessing strong English proficiency. Before verification, the reviewers are familiarized with task-specific guidelines and representative examples covering task definitions, evidence rules, option validity, and common generation errors. Each instance is independently examined by all three reviewers with respect to task suitability, linguistic quality, temporal-evidence support, and answer-label consistency.

The reviewers verify that each rewritten question preserves its intended reasoning objective, that the four candidate options are clear and mutually exclusive, and that the gold answer can be recovered from the corresponding timestamped observations or intervention records. They additionally check that GPT-5.5-based rewriting and option shuffling do not alter the original evidence rule or answer label. Disagreements are resolved through discussion until consensus is reached. Instances with unresolved ambiguity, insufficient evidence, inconsistent source-record alignment, or incorrect labels are removed. The validated instances are then converted into the instruction format used for Stage 3 training.
\section{Details of Experiments}

\subsection{Sensitivity to the Temporal-Token Budget}
\label{apx:token_budget}

We analyze the effect of the temporal-token budget $K$, where $K=|\mathbf{T}|$ denotes the number of tokens produced by the learned-query resampler. We vary $K\in\{8,16,32,64,128\}$ while keeping all other model and training settings unchanged. As shown in Table~\ref{tab:token_budget}, the optimal budget varies across individual tasks, but $K=16$ achieves the highest macro-average accuracy of 49.83. Reducing the budget to 8 lowers the average accuracy to 46.02, whereas increasing it to 32, 64, and 128 yields 47.79, 47.59, and 47.45, respectively. These results show that QA accuracy does not increase monotonically with the number of temporal tokens. We therefore use $K=16$ in all main experiments, as it provides the strongest overall performance while retaining a compact temporal-token interface.

\subsection{Details of Inference Latency}
\label{apx:inference latency}
All latency experiments were conducted on NVIDIA RTX 4090 GPUs with a batch size of 32. We report the average inference time per question, calculated by dividing the total time required to generate complete answers by the number of evaluated samples. The measurement covers the full inference pipeline, including time-series encoding, temporal evidence distillation, and LLM answer generation. Under this setting, ClinPRISM achieves an average latency of 0.15 seconds per question.

\end{document}